\documentclass[11pt,a4paper]{article}
\usepackage[hyperref]{acl2021}
\usepackage{times}
\usepackage{latexsym}

\usepackage{graphicx}
\usepackage{bmpsize}
\newcommand{\bhline}{\noalign{\hrule height 1.5pt}} 
\usepackage{xcolor}
\usepackage{amsmath,amssymb}
\usepackage{multirow}

\usepackage{microtype}

\aclfinalcopy 
\def\aclpaperid{3800} 

\title{Semantic Frame Induction using\\ Masked Word Embeddings and Two-Step Clustering}

\author{
  Kosuke Yamada$^{1}$ \ \ \ \ \ \ \ \ \ Ryohei Sasano$^{1,2}$ \ \ \ \ \ \ \ \ \ Koichi Takeda$^{1}$  \\
  $^{1}$Graduate School of Informatics, Nagoya University, Japan \\
  $^{2}$RIKEN Center for Advanced Intelligence Project, Japan \\
  {\tt yamada.kosuke@c.mbox.nagoya-u.ac.jp}, \\
  {\tt \{sasano,takedasu\}@i.nagoya-u.ac.jp}
}

\date{}

\begin{document}
\maketitle
\begin{abstract}
Recent studies on semantic frame induction show that relatively high performance has been achieved by using clustering-based methods with contextualized word embeddings.
However, there are two potential drawbacks to these methods: one is that they focus too much on the superficial information of the frame-evoking verb and the other is that they tend to divide the instances of the same verb into too many different frame clusters.
To overcome these drawbacks, we propose a semantic frame induction method using masked word embeddings and two-step clustering.
Through experiments on the English FrameNet data, we demonstrate that using the masked word embeddings is effective for avoiding too much reliance on the surface information of frame-evoking verbs and that two-step clustering can improve the number of resulting frame clusters for the instances of the same verb.
\end{abstract}

\section{Introduction}
Semantic frame induction is a task of mapping frame-evoking words, typically verbs, into semantic frames they evoke (and the collection of instances of words to be mapped into the same semantic frame forms a cluster).
For example, in the case of example sentences from FrameNet \cite{baker1998} shown in (1) to (4) in Table \ref{tab:examples}, the goal is to group the examples into three clusters according to the frame that each verb evokes; namely, \{(1)\}, \{(2)\}, and \{(3), (4)\}.
Unsupervised semantic frame induction methods help to automatically build high-coverage frame-semantic resources.

Recent studies have shown the usefulness of contextualized word embeddings such as ELMo \cite{peters2018} and BERT \cite{devlin2019} for semantic frame induction.
For example, the top three methods \cite{arefyev2019, anwar2019, ribeiro2019} in Subtask-A of SemEval-2019 Task 2 \cite{qasemizadeh2019} perform clustering of contextualized word embeddings of frame-evoking verbs.
However, these methods have two potential drawbacks.

\begin{table}[!t]
\centering
\includegraphics[width=\linewidth]{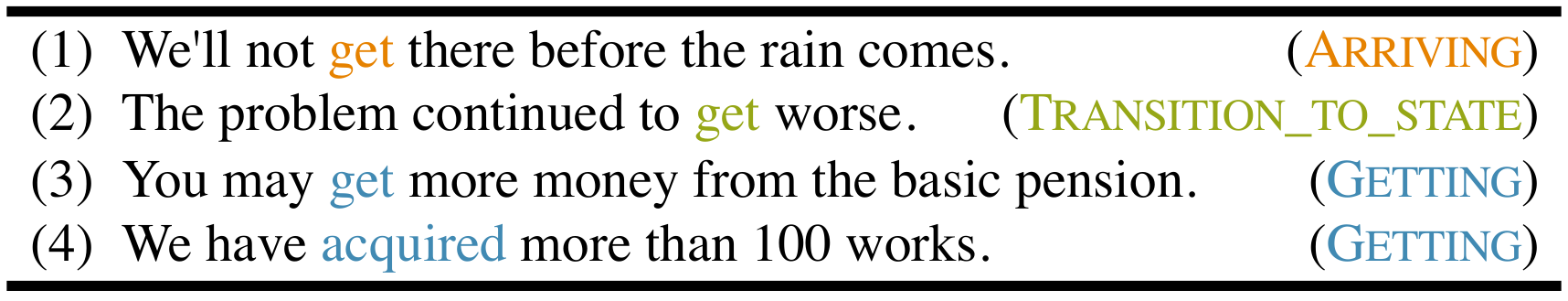}
\caption{\small Example sentences of verbs ``get'' and ``acquire'' and frames that each verb evokes in FrameNet. (\textsc{Frame})}
\label{tab:examples}
\end{table}

\begin{figure}[!t]
\centering
\includegraphics[width=\linewidth]{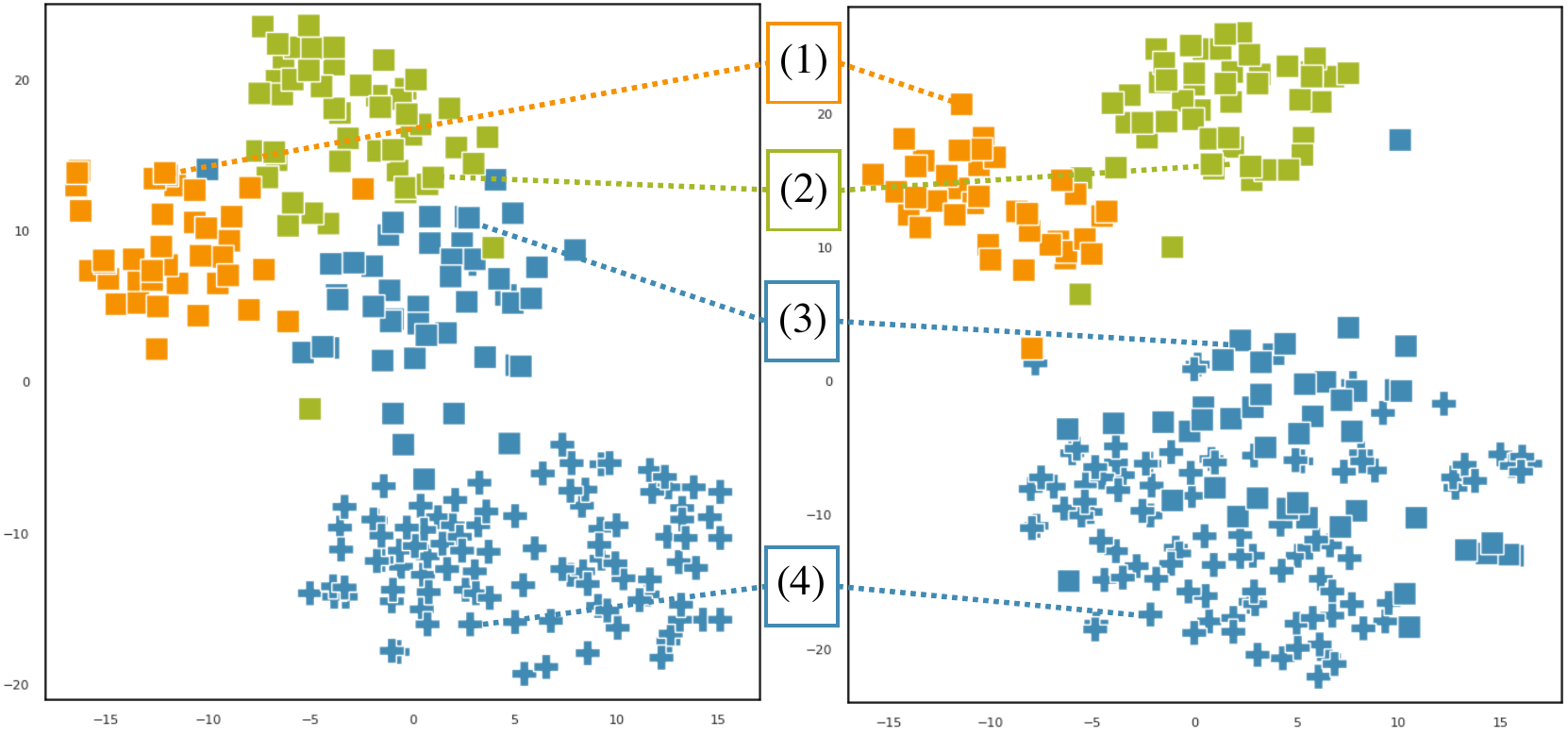}
\caption{\small 2D projections of BERT embeddings of verbs (left) and masked verbs (right). Numbers in the figure correspond to numbers in Table \ref{tab:examples}, {\fontsize{5pt}{0pt} $\blacksquare$} and {\footnotesize\textbf{+}} are verbs ``get'' and ``acquire'', respectively, and each color indicates {\color[HTML]{E68200} \textsc{Arriving}}, {\color[HTML]{96A717} \textsc{Transition\_to\_state}}, and {\color[HTML]{418AB3} \textsc{Getting}} frame.}
\label{fig:visualization}
\end{figure}

First, the contextualized word embeddings of the frame-evoking verbs strongly reflect the superficial information of the verbs.
The left side of Figure \ref{fig:visualization} shows a 2D projection of contextualized embeddings of instances of the verbs ``get'' and ``acquire'' extracted from example sentences in FrameNet.
Specifically, we extracted instances of ``get'' and ``acquire'' from FrameNet, obtained their embeddings by using a pre-trained BERT, and projected them into two dimensions by using t-distributed stochastic neighbor embedding (t-SNE) \cite{maaten2008}.
As shown in the figure, among instances of ``get'', those that evoke the \textsc{Getting} frame tend to be located close to instances of ``acquire'' that evokes the same \textsc{Getting} frame.
However, we can see that the difference between verbs is larger than the difference between the frames that each verb evokes.

To remedy this drawback, we propose a method that uses a masked word embedding, a contextualized embedding of a masked word.
The right side of Figure \ref{fig:visualization} shows a 2D projection of masked word embeddings for instances of the verbs ``get'' and ``acquire''.
The use of masks can hide the superficial information of the verbs, and consequently we can confirm that instances of verbs that evoke the same frame are located close to each other.

The second drawback is that these methods perform clustering instances across all verbs simultaneously.
Such clustering may divide instances of the same verb into too many different frame clusters.
For example, if there are outlier vectors that are not typical for a particular verb, they tend to form individual clusters with instances of other frames in most cases.
To solve this problem, we propose a two-step clustering, which first performs clustering instances of the same verb according to their meaning and then performs further clustering across all verbs.

\section{Proposed Method}
The proposed semantic frame induction method uses masked word embeddings and two-step clustering. 
We explain these details below.

\subsection{Masked Word Embedding}
A masked word embedding is a contextualized embedding of a word in a text where the word is replaced with a special token indicating that it has been masked, i.e., ``[MASK]'' in BERT.
Our method leverages masked word embeddings of frame-evoking verbs in addition to standard contextualized word embeddings of frame-evoking verbs.
In this paper, we consider the following three types of contextualized word embeddings.
\\ \vspace{-3.0ex}
\begin{description}
\setlength{\itemsep}{-0.5ex} 
\item[$v_{\textsc{word}}$:] Standard contextualized embedding of a frame-evoking verb.
\item[$v_{\textsc{mask}}$:] Contextualized embedding of a frame-evoking verb that is masked.
\item[$v_{\textsc{w+m}}$:] The weighted average of the above two, which is defined as:\ \\ \vspace{-3.0ex} 
\begin{equation}
v_{\textsc{w+m}} = (1-\alpha)\cdot v_{\textsc{word}}+\alpha \cdot v_{\textsc{mask}}.
\label{eq:vwm}
\end{equation}
\end{description}

Here, $v_{\textsc{w+m}}$ is the weighted average of contextualized word embeddings with and without masking the frame-evoking verb.
By properly setting the weight $\alpha$ using a development set, we expect to obtain embeddings that properly adjust the weight of superficial information of the target verb and information obtained from its context.
$v_{\textsc{w+m}}$ is identical to $v_{\textsc{word}}$ when $\alpha$ is set to 0 and identical to $v_{\textsc{mask}}$ when $\alpha$ is set to 1.

\subsection{Two-Step Clustering}
\begin{figure}[!t]
\centering
\includegraphics[width=\linewidth]{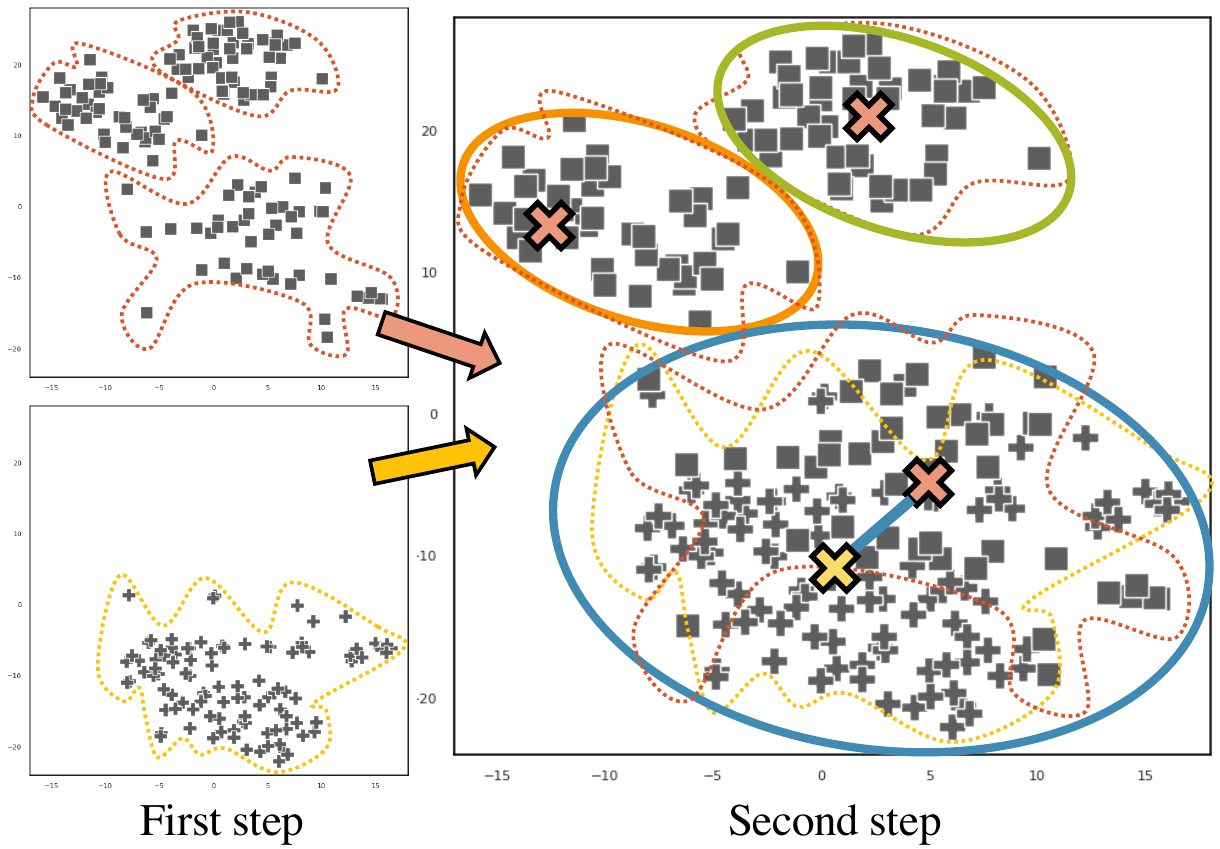}
\caption{\small Flow of the two-step clustering. {\fontsize{5pt}{0pt} $\blacksquare$} and {\footnotesize\textbf{+}} denote the embeddings of ``get'' and ``acquire'', respectively.}
\label{fig:method}
\end{figure}

In the two-step clustering, we first perform clustering instances of the same verb according to the semantic meaning and then perform further clustering across verbs.
Finally, each generated cluster is regarded as an induced frame.
Figure \ref{fig:method} shows the flow of the two-step clustering using the instances of ``get'' and ``acquire'' from FrameNet.
As a result of the clustering in the first step, the instances of ``get'' are grouped into three clusters and the instances of ``acquire'' into one cluster.
In the second step, one of the clusters of ``get'' and the cluster of ``acquire'' are merged. 
Consequently, three clusters are generated as the final clustering result. 
The details of each clustering are as follows.

\paragraph{Clustering Instances of the Same Verb}
The clustering in the first step aims to cluster instances of the same verb according to their semantic meaning.
Since all the targets of the clustering are the same verbs, there should be no difference in the results between the cases using $v_{\textsc{word}}$ and $v_{\textsc{mask}}$ as embeddings.
Therefore, we use only $v_{\textsc{mask}}$ for this process.
We adopt X-means \cite{pelleg2000} or group average clustering based on a Euclidean distance as the clustering algorithm.

While X-means automatically determine the number of clusters, group average clustering requires a clustering termination threshold.
In the group average clustering, the distance between two clusters is defined as the average distances of all instance pairs between clusters, and the cluster pairs with the smallest distance between clusters are merged in order.
The clustering is terminated when there are no more cluster pairs for which the distance between two clusters is less than or equal to a threshold $\theta$.
In this study, $\theta$ is shared across verbs, not determined for each verb.
Note that when $\theta$ is set to a sufficiently large value, the number of clusters is one for all verbs.
To set $\theta$ to an appropriate value, we gradually decrease $\theta$ from a sufficiently large value and fix it to a value where the number of the generated frame clusters is equal to the actual number of frames in the development set.

In the theory of Frame Semantics \cite{fillmore2006} on which FrameNet is based, the association between a word and a semantic frame is called a lexical unit (LU).
Since each cluster generated as the result of clustering in the first step is a set of instances of the same verb used in the same meaning, it can be considered to correspond to an LU.
Therefore, we refer to it as pseudo-LU (pLU). 

\paragraph{Clustering across Verbs}
The clustering in the second step aims to cluster the pLUs generated as the result of the first-step clustering across verbs according to their meaning.
This step calculates average contextualized embeddings of each pLU and then clusters the pLUs by using the calculated embeddings across verbs.
We adopt Ward clustering or group average clustering based on a Euclidean distance as the clustering algorithm.

We need a termination criterion for both clustering algorithms.
A straightforward approach is to use the ratio of the number of frames to the number of verbs.
However, this approach does not work well in this case since there is an upper limit to the number of frame types and the number of frames to be generated does not increase linearly with the number of verbs.
Therefore, in this study, we use the ratio of pLU pairs belonging to the same cluster as the termination criterion.
Specifically, the clustering is terminated when the ratio of pLU pairs belonging to the same cluster $p_{\textsc{f}_1=\textsc{f}_2}$ is greater than or equal to the ratio of LU pairs belonging to the same frame in the development set $p_{\textsc{c}_1=\textsc{c}_2}$.
Here, $p_{\textsc{f}_1=\textsc{f}_2}$ is calculated as:\\\vspace{-2.2ex}
\begin{equation}
p_{\textsc{f}_1=\textsc{f}_2}=\frac
{\mbox{\small \# of pLU pairs in the same cluster}}
{\mbox{\small \# of all pLU pairs}}.
\label{EQ::p}
\end{equation}

While the number of all pLU pairs is constant regardless of clustering process, the number of pLU pairs belonging to the same cluster monotonically increases as the clustering process progresses.
$p_{\textsc{c}_1=\textsc{c}_2}$ can be calculated as well as $p_{\textsc{f}_1=\textsc{f}_2}$ and $p_{\textsc{c}_1=\textsc{c}_2}$ reaches 1 when the number of the entire cluster becomes one cluster.
Therefore, $p_{\textsc{c}_1=\textsc{c}_2}$ is guaranteed to be greater than or equal to $p_{\textsc{f}_1=\textsc{f}_2}$ during the clustering process.
Since the probability that randomly selected LU pairs belong to the same frame is not affected by the data size, the criterion is considered valid regardless of the data size.

\section{Experiment}
We conducted an experiment of semantic frame induction to confirm the efficacy of our method.
In this experiment, the objective is to group the given frame-evoking verbs with their context according to the frames they evoke.

\subsection{Setting}
\paragraph{Dataset}
From Berkeley FrameNet data release 1.7\footnote{\url{https://framenet.icsi.berkeley.edu/}} in English, we extracted verbal LUs with at least 20 example sentences and used their example sentences.
That is, all target verbs in the dataset have at least 20 example sentences for each frame they evoke.
We limited the maximum number of sentence examples for each LU to 100 and if there were more examples, we randomly selected 100.
Note that we did not use the SemEval-2019 Task 2 dataset because the dataset is no longer available as described on the official web page.\footnote{\url{https://competitions.codalab.org/competitions/19159\#learn_the_details-datasets}}

\begin{table}[!t]
\centering
\small
\begin{tabular}{l|rrrr}\bhline
                & \multicolumn{1}{c}{\#Verbs}  & \multicolumn{1}{c}{\#LUs} & \multicolumn{1}{c}{\#Frames} & \multicolumn{1}{c}{\#Examples} \\ \hline
Dev.      &  255  &   300 &      169 & 12,718 \\ 
Test    & 1,017 & 1,188 &      393 & 47,499 \\ \hline
All           & 1,272 & 1,488 &      434 & 60,217 \\ \bhline
\end{tabular}
\caption{\small Statistics of the dataset from FrameNet.}
\label{tab:dataset}
\end{table}

\begin{table*}[t!]
\centering
\small
\begin{tabular}{lccccccc}
\bhline
Model & \multicolumn{2}{c}{Clustering} &   $\alpha$ & \#pLU &      \#C &    \textsc{Pu} / \textsc{iPu} / \textsc{PiF} &   \textsc{BcP} / \textsc{BcR} / \textsc{BcF} \\
\hline
1-cluster-per-head & \multicolumn{2}{c}{1cpv} & -- &-- &   1017 & 88.9 / 39.7 / 54.9 & 86.6 / 33.9 / 48.7 \\ \hline
\newcite{arefyev2019} & \multicolumn{2}{c}{GA (Cosine)} & -- &-- &    995 & 69.9 / 55.1 / 61.6 & 62.8 / 44.0 / 51.7 \\
\newcite{anwar2019} & \multicolumn{2}{c}{GA (Manhattan)} &    -- & -- &    891 & 71.5 / 52.0 / 60.2 & 65.1 / 41.0 / 50.3 \\
\newcite{ribeiro2019} & \multicolumn{2}{c}{Chinese Whispers} &    -- & -- &    542 & 50.9 / 66.3 / 57.5 & 39.4 / 56.7 / 46.5 \\ \hline
One-step & \multicolumn{2}{c}{Ward}               &    0.0 & -- &    393 & 64.3 / 49.5 / 56.0 & 55.2 / 38.9 / 45.6 \\
clustering & \multicolumn{2}{c}{GA}            &   0.0 &   -- &    393 & 38.7 / 64.9 / 48.5 & 26.1 / 52.5 / 34.9 \\\hline\\[-8pt]
& \multicolumn{1}{c}{\footnotesize \textbf{first-step}} 
& \multicolumn{1}{c}{\footnotesize \textbf{second-step}} & & & & \\ 
& 1cpv' & Ward      & 0.8 &  1017 &    164 & 54.8 / 73.1 / 62.7 & 43.1 / 64.3 / 51.6 \\
\multirow{2}{*}{Two-step} & 1cpv' & GA   & 0.9 &  1017 &    412 & 69.0 / 71.3 / 70.1 & 60.5 / 62.3 / 61.4 \\
\multirow{2}{*}{clustering} & GA & Ward & 0.9  &  1196 &    291 & 49.3 / 72.9 / 58.8 & 37.3 / 64.6 / 47.3 \\
 & GA & GA & 0.6 &  1196 &    479 & 63.0 / \textbf{76.3} / 69.0 & 52.8 / \textbf{68.0} / 59.4 \\
& X-means & Ward    & 0.8 &  1043 &    167 & 54.0 / 72.2 / 61.8 & 42.6 / 63.6 / 51.1 \\
& X-means & GA  & 0.7 &  1043 &    410 & \textbf{71.9} / 74.1 / \textbf{73.0} & \textbf{63.2} / 65.5 / \textbf{64.4} \\
\bhline
\end{tabular}
\caption{\small Experimental results. \#pLU denotes the number of pLUs and \#C denotes the number of frame clusters. Note that the actual numbers of LUs and frames are 1,188 and 393, respectively. GA means group average clustering.}
\label{tab:result}
\end{table*}

The extracted dataset contained 1,272 different verbs as frame-evoking words.
We used the examples for 255 verbs (20\%) as the development set and those for the remaining 1,017 verbs (80\%) as the test set.
Thus, there are no overlapping frame-evoking verbs or LUs between the development and test sets, but there is an overlap in the frames evoked.
We divided the development and test sets so that the proportion of verbs that evoke more than one frames would be the same.
The development set was used to determine the alpha of $v_{\textsc{w+m}}$ and the termination criterion for the clustering in each step and layers to be used as contextualized word embeddings.
Table \ref{tab:dataset} lists the statistics of the dataset.

\paragraph{Models}
We compared four models, all combinations of group average clustering or X-means in the first step and Ward clustering or group average clustering in the second step.
We also compared a model that treats all instances of one verb as one cluster (1-cluster-per-verb; 1cpv) and models that treat all instances of one verb as one cluster (1cpv') in the first step and then perform the clustering in the second step.

In addition, we compared our models with the top three models in Subtask-A of SemEval-2019 Task 2.
\newcite{arefyev2019} first perform group average clustering using BERT embeddings of frame-evoking verbs.
Then, they perform clustering to split each cluster into two by using TF-IDF features with paraphrased words.
\newcite{anwar2019} use the concatenation of the embedding of a frame-evoking verb and the average word embedding of all words in a sentence obtained by skip-gram \cite{mikolov2013}.
They perform group average clustering based on Manhattan distance by using the embedding.
\newcite{ribeiro2019} perform graph clustering based on Chinese whispers \cite{biemann2006} by using ELMo embeddings of frame-evoking verbs.

To confirm the usefulness of the two-step clustering, we also compared our models with models that perform a one-step clustering.
For the model, we used Ward clustering or group average clustering as the clustering method and $v_{\textsc{w+m}}$ as the contextualized word embedding.
We gave the oracle number of clusters to these models, i.e., we stopped clustering when the number of human-annotated frames and the number of cluster matched.

\paragraph{Metrics and Embeddings}
We used six evaluation metrics: \textsc{B-cubed Precision} \textsc{(BcP}), \textsc{B-cubed Recall} (\textsc{BcR}), and their harmonic mean, \textsc{F-score} (\textsc{BcF}) \cite{bagga1998}, and \textsc{Purity} (\textsc{Pu}), \textsc{inverse Purity} (\textsc{iPu}), and their harmonic mean, \textsc{F-score} (\textsc{PiF}) \cite{karypis2000}.
We used BERT (bert-base-uncased) in Hugging Face\footnote{\url{https://huggingface.co/transformers/}} as the contextualized word embedding.

\subsection{Results}
Table \ref{tab:result} shows the experimental results.\footnote{The performance of the top three models in Subtask-A of SemEval-2019 Task 2 is lower than reported in the task because the dataset used in this study has a high proportion of verbs that evoke multiple frames and is, therefore, a challenging dataset.
}
When focusing on \textsc{BcF}, which was used to rank the systems in Subtask-A of SemEval-2019 Task 2, our model using X-means as the first step and group average clustering as the second step achieved the highest score of 64.4.
It also got the highest \textsc{PiF} score of 73.0.
The number of human-annotated frames was 393, while the number of generated clusters was 410.
These results demonstrate that the termination criterion of the two-step clustering works effectively.

In all two-step clustering methods, $\alpha$ was tuned between 0.0 and 1.0, which shows that both $v_{\textsc{word}}$ and $v_{\textsc{mask}}$ should be considered.
In addition, $\alpha$ was close to 1.0 for these methods, which indicates that $v_{\textsc{mask}}$ is more useful for clustering instances across verbs.
In contrast, $v_{\textsc{w+m}}$ in the one-step clustering methods was equivalent to $v_{\textsc{word}}$ with $\alpha=0.0$.
This indicates that there is no effect of using $v_{\textsc{mask}}$ for the one-step clustering-based methods.

The two-step clustering-based models that use group average clustering as the second clustering algorithm tended to achieve high scores.
This indicates that the two-step clustering-based approach, which first cluster instances of the same verb and then cluster across verbs, is effective.
However, as to the first clustering, 1cpv' strategy, which treats all the instances of the same verb as one cluster, achieved a higher accuracy than the clustering of the group average method, and achieved an accuracy close to the clustering of X-means, and thus we can say that 1cpv' strategy is effective enough for this dataset.
We think this is due to the fact that the dataset used in this study is quite biased towards verbs that evoke only one frame, and we believe that the effectiveness of the 1cpv' may be limited in a more practical setting.
Further investigation of this is one of our future works.

\section{Conclusion}
We proposed a method that uses masked word embeddings and two-step clustering for semantic frame induction.
The results of experiments using FrameNet data showed that masked word embeddings and two-step clustering are quite effective for this frame induction task.
We will conduct experiments in a setting where nouns and adjectives are also accounted for as frame-evoking words.
The future goal of this research is to build a frame-semantic resource, which requires not only the induction of semantic frames but also the determination of the arguments required by each frame and the induction of semantic roles of the arguments.
A possible extension of our approach is to utilize contextualized word embeddings of arguments of verbs to see if it is possible to generalize our approach for achieving this goal.

\section*{Acknowledgements}
This work was supported by JSPS KAKENHI Grant Numbers 18H03286 and 21K12012.

\bibliographystyle{acl_natbib}
\bibliography{acl2021}
\end{document}